\documentclass{article}

\usepackage{PRIMEarxiv}

\usepackage[utf8]{inputenc} % allow utf-8 input
\usepackage[T1]{fontenc}    % use 8-bit T1 fonts
\usepackage{url}            % simple URL typesetting
\usepackage{booktabs}       % professional-quality tables
\usepackage{amsfonts}       % blackboard math symbols
\usepackage{nicefrac}       % compact symbols for 1/2, etc.
\usepackage{microtype}      % microtypography
\usepackage{lipsum}
\usepackage{fancyhdr}       % header
\usepackage{graphicx}       % graphics
\graphicspath{{media/}}     % organize your images and other figures under media/ folder

\usepackage{times}
\usepackage{epsfig}
\usepackage{amsmath}
\usepackage{amssymb}
\usepackage{multirow}
\usepackage{soul}
\usepackage{float,longtable}
\restylefloat{table}
\usepackage{nccmath} % added to reduce eqn size 
\usepackage[breaklinks=true,bookmarks=false,colorlinks]{hyperref}
\usepackage{wrapfig}

\title{\ Knowledge-Enriched Distributional Model Inversion Attacks
}

\author{
  Si Chen, Mostafa Kahla, Ruoxi Jia \\
  Virginia Tech \\
  Blacksburg, VA \\
  {\tt\small \{chensi, kahla, ruoxijia\}@vt.edu}\\
 \And
 Guo-Jun Qi\thanks{Correspondence to G.-J. Qi, guojunq@gmail.com} \\
Seattle Research Center, Innopeak Technology \\
Bellevue, WA\\
{\tt\small guojun.qi@innopeaktech.com}
}

\begin{document}
\maketitle

\begin{abstract}
    Model inversion (MI) attacks are aimed at reconstructing training data from model parameters. Such attacks have triggered increasing concerns about privacy, especially given a growing number of online model repositories. However, existing MI attacks against deep neural networks (DNNs) have large room for performance improvement. 
   %A natural question is whether the underperformance is because the target model does not memorize much about its training data or it is simply an artifact of imperfect attack algorithm design? 
   %This paper shows that it is the latter. 
   We present a novel inversion-specific GAN that can better distill knowledge useful for performing attacks on private models from public data. In particular, we train the discriminator to differentiate not only the real and fake samples but the soft-labels provided by the target model. 
   %We also introduce a loss term to the generator training that favors the generated images that can prompt the target model to make high-confidence prediction. 
   Moreover, unlike previous work that directly searches for a single data point to represent a target class, we propose to model a private data distribution for each target class. Our experiments show that the combination of these techniques can significantly boost the success rate of the state-of-the-art MI attacks by $150\%$, and generalize better to a variety of datasets and models. Our code is available at \url{https://github.com/SCccc21/Knowledge-Enriched-DMI}.
\end{abstract}

% keywords can be removed
% \keywords{First keyword \and Second keyword \and More}

\section{Introduction}
Many attractive applications of machine learning techniques involve training models on sensitive and proprietary datasets. One major concern for these applications is that models could be subject to privacy attacks and reveal inappropriate details of the training data. One type of privacy attacks is MI attacks, aimed at recovering training data from the access to a model. The access could either be black-box or white-box. In the blackbox setting, the attacker can only make prediction queries to the model, while in the whitebox setting, the attacker has complete knowledge of the model. Given a growing number of online platforms where users can download entire models, such as Tensorflow Hub\footnote{\url{https://www.tensorflow.org/hub}} and ModelDepot\footnote{\url{https://modeldepot.io/}}, whitebox MI attacks have posed an increasingly serious threat to privacy.

Effective MI attacks have been mostly demonstrated on simple models, such as linear models, and low-dimensional feature space~\cite{fredrikson2014privacy,fredrikson2015model}. MI attacks are typically cast as an optimization problem that seeks for the most likely input examples corresponding to a target label under the private model. When the target model is a DNN, the underlying attack optimization problem becomes intractable and solving it via gradient methods in an unconstrained manner will end in a local minimum that contains meaningless pixel patterns. Previous MI attack models like ~\cite{zhang2020secret} explore the idea of distilling a generic prior from potential public data via a GAN generator and using it to guide the inversion process. For instance, to attack a face recognition classifier trained on private face images, one can train a GAN with public face datasets  to learn generic statistics of real face images 
%like how a normal face might be structured 
and then solving the attack optimization over the latent space of the GAN rather than in an unconstrained ambient space. %So far, this idea has led to the state-of-the-art performance for attacking DNNs. 

However, there still exists a large room to improve the attack performance. For instance, the top-one identification accuracy of face images inverted from the state-of-the-art face recognition classifier is 45\%.
%and further decreases when the public data has a large distributional shift from the private data. 
A natural question is: \emph{Is the underperformance of MI attacks against DNNs because DNNs do not memorize much about private data or it is simply an artifact of imperfect attack algorithm design?} This paper shows that it is the latter.

We reveal a variety of drawbacks associated with the the current MI attacks against DNNs. Particularly, we notice that the previous state-of-the-art approach suffers from the two key limitations: 1) The information about private classifier is not sufficiently explored for distilling knowledge from public data. Previous works ignore the important role of the target classifier in adapting the knowledge distilled from the public data for training the MI attack model on the target classifier. Indeed, given a target classifier to attack, we can also use its output labels to distill which public data are more useful in inverting the target model to recover the private training examples of the given labels.   
2) Prior works made a simplified one-to-one assumption in recovering a single example for a given label of the target model. However, in real scenarios, inverting a model should naturally result in a distribution of training examples corresponding to the given label. This inspires us to recover a data distribution in the MI attack in line with such a many-to-one assumption.

%uses random initialization to solve the attack optimization problem and may still get stuck in ``bad'' local minima.

To address the first limitation, we propose to tailor the training objective of the GAN to the inversion task. Specifically, for the discriminator, we propose to leverage the target model to label the public dataset and train the discriminator to differentiate not only the real and fake samples but also the labels. This new training scheme will force the generator to retain image statistics that are more relevant to infer the classes of the target model, which are likely to occur in the unknown private training data. 
%For the generator, we introduce a loss term that favors the images with high prediction confidence when fed into the target model. This loss term will also force the learned embedding to be close to training data as the training data will also achieve high confidence when fed into the target model. 
To overcome the second limitation, we propose to explicitly parameterize the private data distribution and solve the attack optimization over the distributional parameters. Moreover, this will lead us to explore a distribution in which each point with large probability mass will achieve a good attack performance. 
%The distributional MI attack  allows us to circumvent the large variance of performance due to random initialization in the previous work. 
We perform experiments on various datasets and network architectures and show that such a distributional MI attack by distilling public-domain knowledge tailored for private labels can significantly improve the previous state-of-the-art attack against DNNs, even when the public data have no overlap with the private labels of the target network. 
%We also introduce a new threat model in which the attacker, unlike previous work on MI attacks, can utilize the availability of multiple models trained on the same private dataset to perform an attack that achieves the highest reported results.

%a large distributional shift from the private data.
The paper is organized as follows. In Section 2 we introduce the related work on model inversion attacks. In Section 3 we describe our proposed inversion-specific GAN and distributional recovery.
%and introduce the new threat model.
We assess the performance of the proposed method in Section 4 and show the extend application of the method to multi-model inversion attacks. Finally, we conclude and discuss our key findings in Section 5.

%%%%%%%%%%%%%%%%%%%%%%%%%%%%%%%%%%%%%%%%%%%%%%%%%%%%
%%%%%%%%%%%%%%%% Related work %%%%%%%%%%%%%%%%%%%%%%
%%%%%%%%%%%%%%%%%%%%%%%%%%%%%%%%%%%%%%%%%%%%%%%%%%%%
\section{Related work}
The general goal of privacy attacks against machine learning models is to gain knowledge which is not intended to be shared, such as knowledge about the training data and information about the model. Attacks can be categorized into four types according to the specific goals: model extraction, membership inference, property inference, and model inversion. Model extraction attacks~\cite{merity2016pointer, krishna2019thieves, orekondy2019knockoff, correia2018copycat} try to create a substitute model that learns the same task as the target model while performing equally good or even better. The other three kinds of attacks focus on exposing secrets about training data: membership inference attacks~\cite{shokri2017membership} try to determine whether a given data point is used as part of the training set; property inference attacks~\cite{ateniese2015hacking, ganju2018property, melis2019exploiting} try to extract dataset properties which are not explicitly correlated to the learning task (e.g., extracting the ratio of women and men in a patient dataset where this information is unlabeled). The goal of MI attacks is to recreate training data or sensitive attributes. 

The first MI attack algorithm was proposed in~\cite{fredrikson2014privacy}, which follows the Maximum a Posterior (MAP) principle and constructs the input features that maximize the likelihood of observing a given model response and other possible auxiliary information. The authors applied the algorithm to attack a linear regression model that predicts medical dosage and showed that the algorithm can successfully invert genetic markers which are used as part of the input features.

Fredrikson \textit{et al.}~\cite{fredrikson2015model} applied the MAP attack idea to more complex models, including decision trees and shallow neural networks. Specifically, for neural networks with high-dimensional input features, the authors proposed to utilize gradient descent to solve the underlying attack optimization problem. Although the algorithm significantly outperforms random guessing when tested on some shallow networks and single-channel images, the reconstructions are blurry and can hardly reveal private information. Besides, the algorithm completely fails when tested on DNNs and three-channel images. 

To improve the attack performance for DNNs with high-dimensional input, a two-pronged attack approach \cite{zhang2020secret} was proposed which trains a GAN on public data (which could have no class intersection with private data and no labels), and then uses the GAN to search for the real examples that maximize the response to given classes. However, the resultant GAN fails to distill the private knowledge customized for the specific classes of interest in the target network, and the associated MI attack cannot recover the distribution of examples corresponding to those private classes.

The aforementioned works for attacking neural nets focused on the white-box setting and attacking a single model that is learned offline. Recent work has also looked into other attacker models. For instance, Yang \textit{el al.}~\cite{yang2019adversarial} studied the blackbox attack and proposed to train a separate model that swaps the input and output of the target model to perform MI attacks. Salem \textit{et al.}~\cite{salem2019updates} studied the blackbox MI attacks for online learning, where the attacker has access to the versions of the target model before and after an online update and the goal is to recover the training data used to perform the update. 

Moreover, the algorithms of MI attacks resemble an orthogonal line of work on feature visualization~\cite{nguyen2016synthesizing,yosinski2015understanding}, which also attempts to reconstruct an image that maximally activates a target network. The proposed work differs from these existing works on feature visualization in that our algorithm  customizes the public-to-private knowledge distillation to train the GAN and a novel formulation is presented for data distribution synthesis which results in more realistic image recovery.

\section{The Proposed Approach}

\subsection{Overview of our attack}
\paragraph{Attack model} 
This paper focuses on the whitebox MI attack, in which the attacker has complete access to the target network $T$. The goal of the attacker is to discover \emph{a representative input feature} $x$ associated with a specific label $y$. We will use face recognition as a running example for the target network. Face recognition classifiers label an image containing a face with the label corresponding to the identity depicted in the image. The corresponding attack goal is to recover a representative face image for any given identity based on the target classifier parameters.

Existing MI attacks boil down to synthesizing the most likely input for the target network. Specifically, the following optimization problem is solved to synthesize the input for a given label $y$: $\max_x \log T_y(x)$, where $T_y(x)$ is the probability of label $y$ output by the model $T$ given the input $x$. When $T$ is a DNN and $x$ is high-dimensional (e.g., images), the corresponding optimization becomes nonconvex and performing gradient descent easily gets stuck in local minima. The local minima might not be semantically meaningful at all. For instance, when the model input is an image, such local minima could be meaningless patterns of pixels. 

\begin{figure*}[t!]
\centering
\includegraphics[scale=0.34]{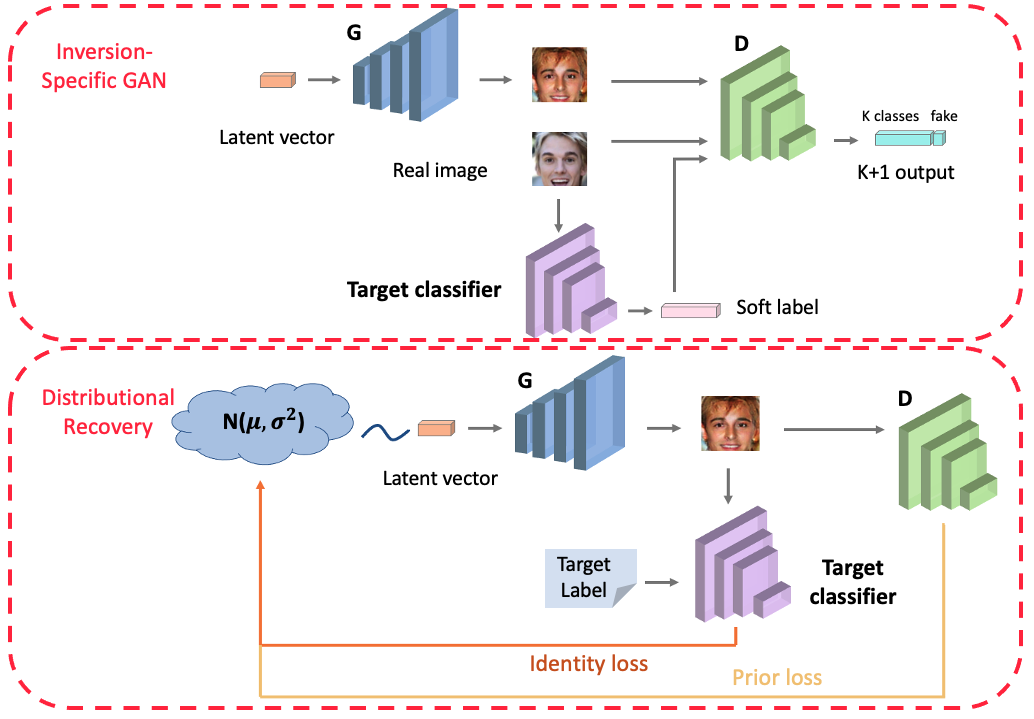}
\caption{Overall architecture of the proposed attack algorithm. \textit{Step 1.} Build an inversion-specific GAN to distill private information. \textit{Step 2.} Recover the distribution of private domain. Note that both the generator and discriminator are fixed at Step 2.}
\label{fig: overall}
\end{figure*}

%Inspired by~\cite{zhang2020secret}, 
The proposed proposed attack algorithm consists of two steps. The first step is to train a GAN having knowledge about the private classes of the target model from public data. Instead of training a generic GAN, we customize the training objective for both generator and discriminator so as to better distill the private-domain information about the target model from public data. In the second step, we make use of the generator learned in the first step to estimate the parameters of the private data distribution. The overall architecture of our method is shown in Figure~\ref{fig: overall}.

\subsection{Building an Inversion-Specific GAN} 

To distill the useful knowledge about the target model from public data, we propose to adopt a discriminator that is not only able to differentiate real data from the fake, but also to discriminate between the class labels under the target network. 

Suppose that the target network classifies a sample into one of $K$ possible classes. Our discriminator $D$ is a $(K+1)$-classifier \cite{salimans2016improved}, where the first $K$ classes correspond to the labels of the target network and the $(K+1)$-th class represents fake samples. To train such a discriminator, we use the target network $T$ to generate a soft label $T(x)$ for each image from the public dataset. 

Formally, the training loss for $D$ has two parts:
\begin{align}
   L_D = L_\text{supervised} + L_\text{unsupervised} 
\end{align}
% \[L_{supervised} = CrossEntropy(D(x), T(x)))\]
where
%\[L_{supervised}=-\mathbb{E}_{x, y \sim p_{{data }}(x, y)} \log p_{{disc}}(y \mid x, y<K+1)\]
\begin{align}
    L_\text{supervised}=-\mathbb{E}_{x \sim p_{\text{data }}(x)} \sum_{k=1}^K T_k(x) \log p_{\text{disc}}(y=k \mid x)
\end{align}
% \[L_{supervised}=-\mathbb{E}_{x \sim p_{{data }}(x)} \sum_{k=1}^K T_k(x) \log p_{{disc}}(y=k \mid x)\]
and
\begin{align}
   L_\text{unsupervised}=-\{&\mathbb{E}_{x \sim p_{\text{data }}(x)} \log D(x)+ \\
   &\mathbb{E}_{z \sim \text{noise}} \log (1-D(G(z)))\}.
\end{align}
Here $p_\text{data}$ is the distribution of public data, and $p_\text{disc}(y|x)$ is the probability that the discriminator predicts $x$ as class $y$. The random noise $z$ is sampled from $\mathcal{N}(0,I)$, and $T_k(x)$ is the $k$-th dimension of the soft label produced by the target network. The discriminator $D(x)$ outputs the probability of $x$ being a real sample, and therefore we have $D(x)\triangleq p_\text{disc}(y<K+1|x)$.

Intuitively, using the public data with soft-labels to train the discriminator encourages the generator to produce image statistics that help predict the output classes of the target model. Such image statistics are also likely to be present in the private training data. Hence, the proposed training process can potentially guide the generator to produce images that share more common characteristics with the private training data.

For training the generator, we adopt the following feature-matching 
loss~\cite{salimans2016improved} to align the generated images with the real counterparts based on the learned features $\mathbf{f}(x)$  encoded in an intermediate layer of the discriminator:
%introduce an entropy regularizer into the canonical feature matching-based training objective~\cite{salimans2016improved}:
\begin{align}
    L_G = \left\|\mathbb{E}_{x \sim p_\text{data}} \mathbf{f}(x)-\mathbb{E}_{z \sim \text{noise
    }} \mathbf{f}(G(z))\right\|_{2}^{2} + \lambda_h L_\text{entropy} 
\end{align}
where $L_\text{entropy}$ is an entropy regularizer \cite{grandvalet2005semi}.

%$\mathbf{f}(x)$ denotes activation on an intermediate layer of the discriminator.
% and
% \begin{align}
%     L_\text{entropy} &= \mathbb H(p_\text{disc}(1\leq y\leq K|G(z)))\\
% &=-\sum_{k=1}^K p_\text{disc}(y=k|G(z))\log p_\text{disc}(y=k|G(z)).
% \end{align}

The intuition of the entropy regularization term is simple. Because the target network is trained on the private data, the private data should have high confidence when fed into the target network and in turn should get low prediction entropy. In order to encourage the data distribution learned from public data to mimic the private data, we explicitly constrain the entropy in the loss function so that the generated data will have low entropy under the target network.

\subsection{Distributional Recovery} 
Given the GAN trained above on the public data under the guidance of the target network, the second step of the MI attack tries to find the private data which achieves the maximum likelihood under the target classification network while containing realistic images. While existing works focus on generating a representative image of a given identity, there ought to be a variety of training examples corresponding to one identity -- indeed, the classifier is a many-to-one 
mapping. To this end, we propose to recover a data distribution instead of a single point to invert the target model for a given label $k$ of identity.
%distributional recover which aims to find the distribution for each identity.

Specifically, given an identity label $k$, we model the private data distribution by $G(z')$, where $G$ is the generator trained in the first step and $z'$ is sampled from $p_\text{gen}=\mathcal{N}(\mu,{\sigma}^2)$ with two {\em learnable} parameters $\mu$ and $\sigma^2$. We then minimize the following objective function to generate the samples for the given class $k$ from the private classifier $T$ by estimating $\mu$ and $\sigma$: 
\begin{align}
    L = L_\text{prior}+ \lambda_i L_\text{id}
\end{align}
where $\lambda_i$ is a positive balancing hyperparameter, and
\begin{align}
    &L_\text{prior} = -\mathbb{E}_{z'\sim p_\text{gen}}\log D(G(z'))\\
%   & L_\text{id} = \mathbb H(T(G(z)), y=k) = -\mathbb{E}_{z\sim p_\text{gen}} \log{T_k(G(z))}
    &L_\text{id} = -\mathbb{E}_{z'\sim p_\text{gen}}T_k(G(z'))
\end{align}
Here the prior loss $L_\text{prior}$ penalizes unrealistic images and the identity loss $L_\text{id}$ encourages the estimated private data distribution
to have high likelihood of 
% the generated samples $G(z),z\sim p_{\text{gen}}$ 
being assigned to the given target label $k$ under the targeted network $T$.

To estimate $\mu$ and $\sigma^2$ directly through the back-propagation, we adopt the reparameterization trick~\cite{kingma2013auto} to make $L_\text{prior}$ and $L_\text{id}$ differentiable:
\begin{align}
    z' = \sigma\epsilon+\mu, \epsilon\sim \mathcal{N}(0,I)
\end{align}
We can now form Monte Carlo estimates of expectations of $L_\text{prior}$ and $L_\text{id}$ as follows and optimize them with respect to $\sigma$ and $\mu$:
\begin{align}
    &L_\text{prior} = -\frac{1}{L}\sum_{l=1}^L\log D(G(\sigma\epsilon_l+\mu))\\
   & L_\text{id} =- \frac{1}{L}\sum_{l=1}^L\log{T_k(G(\sigma\epsilon_l+\mu))} 
\end{align}
where $\epsilon_l\sim \mathcal{N}(0,I)$ for $l=1,\ldots,L$.

\begin{figure}[t!]
\centering
\includegraphics[scale=0.35]{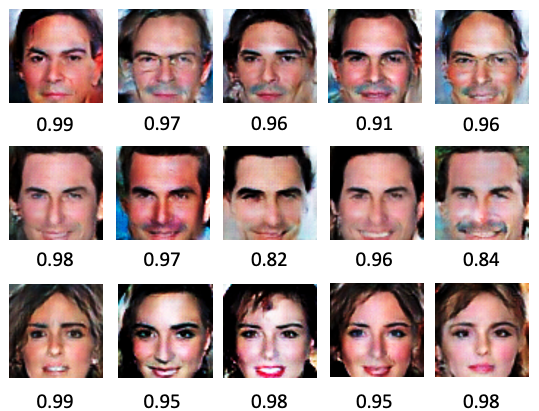}
\caption{Examples of images obtained by inverting a target face recognition model. Each row corresponds to an identity. The numbers beneath each image show high softmax scores for the corresponding identity by the evaluation classifier, demonstrating these generated images successfully attack the target model by exposing its private information.}
\label{fig:diverse}
\end{figure}

Once $\mu$ and $\sigma$ are estimated, the distribution of the learned training examples corresponding to the label $k$ is given implicitly by sampling from $G(z')$ with $z'\sim \mathcal N(\mu, \sigma^2)$. Figure \ref{fig:diverse} shows some examples obtained by sampling from $G(z')$. These examples show a variety of face images are obtained for each identity by inverting a target face recognition model, containing variations in face poses, expressions, hairs and beards.  This suggests that a natural many (faces)-to-one (identity) mapping is learned through a MI attack. We can also model the distribution by multi-variant Gaussian to have further improvement. And this will be left for future work.

%%%%%%%%%%%%%%%%%%%%%%%%%%%%%%%%%%%%%%%%%%%%%%%%%%%%
%%%%%%%%%%%%%%%%%% Experiment %%%%%%%%%%%%%%%%%%%%%%
%%%%%%%%%%%%%%%%%%%%%%%%%%%%%%%%%%%%%%%%%%%%%%%%%%%%
\section{Experiment}
In this section, we will evaluate our proposed attack in terms of the performance to recover a representative input from a target model. The baseline that we will compare against is the generative MI attack (GMI) proposed in~\cite{zhang2020secret}, which achieved the state-of-the-art result for attacking DNNs.

\subsection{Experimental setting}
\paragraph{Dataset.} We study attacks against models built for different prediction tasks, including face recognition, digit classification, object classification, and disease prediction. For face recognition, we use (1) the CelebFaces Attributes Dataset~\cite{liu2015deep} (CelebA) containing 202,599 face images of 10,177 identities with coarse alignment, (2) Flickr-Faces-HQ (FFHQ) Dataset  containing 70,000 high-quality images with considerable variation in terms of age, ethnicity and image background, and (3) FaceScrub consisting of 106,863 face images of male and female 530 celebrities, with about 200 images per person. We use aligned versions of above face datasets, and crop the images at the center and resize them to $64\times64$ so as to remove most background. For digit classification, we use the MNIST handwritten digit data~\cite{Lecun1998Gradient-Based}. For object classification, we adopt the CIFAR-10 dataset~\cite{cifar10}. For disease prediction, we use the Chest X-ray Database~\cite{wang2017chestx} (ChestX-ray8).

\paragraph{Models.} Following the settings in~\cite{zhang2020secret}, we implement several different target networks with varied complexities. Some of the networks are adapted from existing ones by adjusting the number of outputs of their last fully connected layer to our tasks. For the face recognition task, we use three different network architectures: (1) VGG16 adapted from \cite{simonyan2014very}; (2) ResNet-152 adapted from \cite{he2016deep}; (3) face.evoLve adpated from \cite{cheng2017know}. For digit classification on MNIST, we use a network which consists of 3 convolutional layers and 2 pooling layers. For object classification, we use VGG16. For the disease prediction on ChestX-ray8, we use Resnet-18 adapted from \cite{he2016deep}.

\paragraph{Attack Implementation.} We split each dataset into two disjoint parts: one part used as the private dataset to train the target network and the other as a public dataset. \emph{The public data, throughout the experiments, do not have class intersection with the private training data of the target network}. Therefore, the public dataset in our experiment only helps the adversary to gain knowledge about features generic to all classes and does not provide information about private, class-specific features for training the target network. For CelebA, we use 30,027 images of 1000 identities as private set and randomly choose 30,000 images of other identities as public set to train GAN. For MNIST and CIFAR10, we use all of the images with label 0, 1, 2, 3, 4 as private set and rest images with label 5, 6, 7, 8, 9 as public set. For ChestX-ray8, we use 10,000 images with label "Atelectasis", "Cardiomegaly",  "Effusion", "Infiltration", "Mass", "Nodule", "Pneumonia" as private set and 10,000 images belongs to other 7 classes as public set. We train the target networks using the SGD optimizer with the learning rate $10^{-2}$, batch size $64$, momentum $0.9$ and weight decay $10^{-4}$. For training GANs, we use the Adam optimizer with the learning rate $0.004$, batch size $64$, $\beta_1=0.5$ and $\beta_2=0.999$ as~\cite{kingma2014adam}. The weight for entropy regularization term is $\lambda_h = 1\mathrm{e}{-4}$. For the step of distributional recovery, we set $\lambda_i=100$; the distribution is initialized with $\mu=0, \sigma=1$ and optimized for 1500 iterations.

\paragraph{Evaluation Protocol.} For our proposed attack, we draw $5$ random samples of $\epsilon$ and generate corresponding images $G(\sigma\epsilon+\mu)$. For the baseline attack, we re-start the attack for $5$ times with random initialization. To evaluate the reconstruction of a representative input, we compute the average of attack performance on the $5$ reconstructed images. 

\paragraph{Evaluation Metrics.} Evaluating the MI attack performance requires gauging the amount of private information about a target label leaked through the synthesized images. We conduct both qualitative evaluation through visual inspection as well as quantitative evaluation. The quantitative metrics that we use to evaluate the attack performance largely follow the existing literature~\cite{zhang2020secret}, including attack accuracy and K-nearest neighbor feature distance. They are generally aimed at measuring the semantic similarity between private data and reconstructions. In addition, we incorporate a metric for image quality, namely, Fréchet Inception Distance (FID)~\cite{heusel2017gans}, as part of our evaluation. The metrics are expounded as follows.
\begin{itemize}
    \item \underline{Attack Accuracy (Attack Acc)}. We build an \emph{evaluation classifier} that predicts the identity based on the input reconstructed image. If the evaluation classifier achieves high accuracy, the reconstructed image is considered to expose private information about the target label. It is shown in~\cite{zhang2020secret} that the reconstructed images may overfit the target network; in other words, reconstructed images could be meaningless pixel patterns but achieve high prediction accuracy when evaluated with the target network. Hence, the evaluation classifier should be different from the target network. Moreover, the evaluation classifier should achieve high performance, because we are using it as a proxy for a human observer or an oracle to judge whether a reconstruction captures sensitive information. The attack accuracy is measured by the prediction accuracy of the evaluation classifier on reconstructed images. 
    % is measured by a evaluation classifier which predicted labels for generated attack samples. To make sure the evaluation is accurate, we chose highly performant architecture which is also different from the target network. 
    For all the face image datasets, we use the model in~\cite{cheng2017know} as our evaluation classifier, which is pretrained on MS-Celeb-1M~\cite{guo2016ms} and fine-tuned on the training set of the target networks. For MNIST, we train a new evaluation classifier which consists of 5 convolutional layers and 2 pooling layers on all of the 10 digits. For ChestX-ray8, the evaluation classifier is adapted from~\cite{simonyan2014very}. For CIFAR10, we use ResNet-18 adapted from~\cite{he2016deep}.
    
     \item \underline{K-Nearest Neighbor Distance (KNN Dist)}. KNN Dist is the shortest feature distance from a reconstructed image to the real private training images for a given class. The feature distance is measured by the $l_2$ distance between two images when projected onto the feature space, i.e.,
     the output of the penultimate layer of the evaluation classifier.
     
    \item \underline{FID}. FID score measures feature distances between real and fake images, and lower FID values indicate better image quality and diversity. We found that reconstructed images which the evaluation classifier predicts into the target label tend to achieve lower FID scores. Hence, the FID score and attack accuracy are correlated with one another. To make FID a complementary metric to attack accuracy, we only calculate the FID score of those reconstructions which are successfully recognized as the target class by the evaluation classifier. The idea of this FID score is to measure how much more detailed information is leaked from a reconstruction that can successfully recover the semantics.
    
\end{itemize}

\subsection{Result}
\paragraph{Comparison with previous state-of-the-art.} We compare our attack with the baseline for attacking various models built on the same dataset, namely, CelebA. The models include VGG16, ResNet152, and face.evolve, which have increased complexity. Among these models, face.evolve achieves state-of-the-art face recognition performance. The results for attacking these models are shown in Table~\ref{tab: model}, showing that our approach significantly improves the GMI on all the target models. Notably, our approach also enjoys lower performance variance across different target identities compared with the GMI.

\begin{table*}[ht!]
\centering
\resizebox{0.9\textwidth}{!}{
\begin{tabular}{ccccccc}
\toprule
            & \multicolumn{2}{c}{\textbf{face.evolve}}                         & \multicolumn{2}{c}{\textbf{IR152}}   &  \multicolumn{2}{c}{\textbf{VGG16}}                             \\ 
\textbf{}     &\textbf{GMI}               & \textbf{Ours}                       & \textbf{GMI}               & \textbf{Ours}                       & \textbf{GMI}               & \textbf{Ours}                       \\ \midrule
\textbf{Attack Acc $\uparrow$} & .31$\pm$.0039 & \textbf{.81$\pm$.0016} & .32$\pm$.0027 & \textbf{.81$\pm$.0015} &.21$\pm$.0020  & \textbf{.72$\pm$.0018}\\
\textbf{Top-5 Attack acc $\uparrow$}  & .53$\pm$.0015 & \textbf{.96$\pm$.0004} & .57$\pm$.0005 & \textbf{.96$\pm$.0001}&.43$\pm$.0014 &\textbf{.92$\pm$.0003} \\
\textbf{KNN Dist $\downarrow$} &1703.52 &\textbf{1358.23} &1673.05  &\textbf{1324.72}  &1772.50 &\textbf{1380.22} \\
\textbf{FID $\downarrow$} &33.81 &\textbf{25.28} &50.11  &\textbf{26.35}  &52.51 &\textbf{23.72} \\
\bottomrule
\end{tabular}
}
\caption{Attack performance comparison on various models trained on CelebA. $\uparrow$ and $\downarrow$ respectively symbolize that higher and lower scores give better attack performance.}
\label{tab: model}
\end{table*}

\begin{figure*}[ht!]
\centering
\includegraphics[scale=0.40]{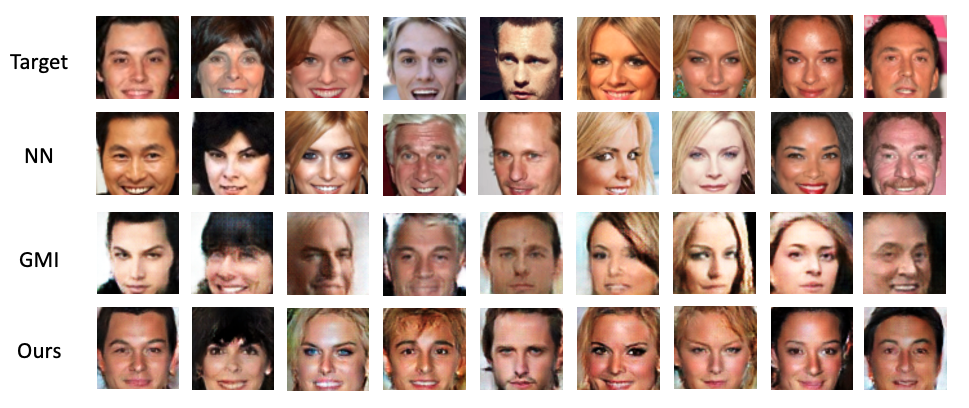}
\caption{Qualitative comparison for attacking a face recognition model trained on CelebA. The first row shows ground truth images for target identities. The second row shows nearest neighbors of the target images from public domain. And the third and last rows demonstrate the reconstructions produced by the GMI attack and our attack, respectively.}
\label{fig:attack}
\end{figure*}

The performance improvement achieved by our attack is further corroborated by Figure~\ref{fig:attack}, which exhibits ground truth private images and corresponding reconstructions given by our attack and the GMI. We can see that our reconstructions can mostly better preserve the facial features of a given identity than the baseline. Since both our approach and the GMI are based on a GAN trained over public data, a natural question is whether these two approaches simply memorize the public data and output a public examples similar to the target identity? To answer this question, we also exhibit the nearest neighbors in the public dataset for each of the target images in Figure~\ref{fig:attack}. We calculate the nearest neighbors based on the distance between deep feature representations extracted from the evaluation classifier in order to capture the perceptual similarity between two images~\cite{zhang2018unreasonable}. Comparison between the nearest neighbors and our generated samples shows that both GMI and our approach do not simply ``memorize'' the similar images in the public domain; instead, they attempt to synthesize new images that expose sensitive attributes while remaining realistic.

Moreover, we examine the performance of the proposed
attack for recovering some implicit attributes of the private images, such as gender, age, hair style, among others. Table~\ref{tab: attributes} shows that our attack also outperforms GMI in terms of recovering the implicit attributes.

\begin{table}[H]
\centering
\resizebox{0.35\columnwidth}{!}{
\begin{tabular}{ccc}
\toprule
\multirow{2}{*}{\textbf{Attributes}} &\multicolumn{2}{c}{\textbf{Attack Acc $\uparrow$}}\\
& \textbf{GMI} & \textbf{Ours}                      \\ \midrule
\textbf{Blond Hair}                  & 84                                                   & \textbf{85}                               \\
\textbf{Bushy Eyebrows}              & \textbf{85}                                                 & \textbf{85}                               \\
\textbf{Glasses}                     & 95                                                   & \textbf{96}                               \\
\textbf{Male}                        & 86                                                   & \textbf{94}                               \\
\textbf{Mustache}                    & 90                                                   & \textbf{93}                               \\
\textbf{Young}                       & 72                                                   & \textbf{82}                               \\
\textbf{5 o'clock shadow}         & 83                                                   & \textbf{87}                               \\
\textbf{Arched Eyebrows}            & 65                                                   & \textbf{70}                               \\
\textbf{Big Nose}                   & 73                                                   & \textbf{78}                               \\
\textbf{Heavy Makeup}               & 61                                                   & \textbf{72}                               \\
\textbf{Narrow Eyes}                & 78                                                   & \textbf{82}                               \\
\textbf{No Beard}                   & 84                                                   & \textbf{90}                               \\
\textbf{Wearing Lipstick}           & 57                                                   & \textbf{74}                               \\
\bottomrule
\end{tabular}
}
\caption{Comparison of implicit attributes recovering between GMI and our proposed method. Attack accuracy is measured by an attributes classifier trained on CelebA. }
\label{tab: attributes}
\end{table}

\begin{figure}[t!]
\centering
\includegraphics[width=.4\textwidth]{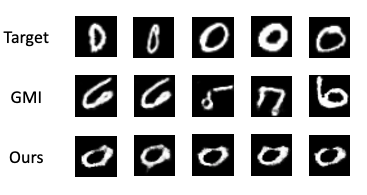}
\caption{MNIST samples generated by GMI and our method when attacking digit ``0".}
\label{fig:mnist}
\end{figure}

Table~\ref{tab: dataset} compares the attack performance of our attack and the GMI on various datasets. We can see that our method outperforms the GMI by a large margin. One interesting finding is that, when attacking digit recognition model trained on MNIST, GMI generates images that can be successfully recognized as the target digits by the target classifier but cannot be predicted into the target digits by the evaluation classifier and the average attack accuracy is close to 0. As shown in Figure~\ref{fig:mnist}, when attacking digit ``0,'' GMI tends to generate ``6"  because it only sees ``6'' in the public data. However, the generated samples can achieve high prediction accuracy under the target network, because it is trained to only predict 0-4, while having low prediction accuracy under the evaluation classifier which can predict all ten digits. In contrast, our attack can successfully reconstruct ``0'' even though it also only sees 5-9 in the public data. This demonstrates that our customized training of GAN can indeed help retain those features in the public data that are more likely to appear in private data.
% Our method can do beyond ``memorizing'' and hence achieves better performance. 

\begin{table*}[ht!]
\centering
\resizebox{\textwidth}{!}{
\begin{tabular}{ccccccccc}
\toprule
       &\multicolumn{2}{c}{\textbf{CelebA}}     & \multicolumn{2}{c}{\textbf{MNIST}}                 & \multicolumn{2}{c}{\textbf{ChestX-ray8}}                         & \multicolumn{2}{c}{\textbf{CIFAR10}}                             \\ 
\textbf{}  & \textbf{GMI} & \textbf{Ours}   & \textbf{GMI} & \textbf{Ours}                       & \textbf{GMI}               & \textbf{Ours}                       & \textbf{GMI}               & \textbf{Ours}                       \\ \midrule
\textbf{Attack Acc $\uparrow$} &.21$\pm$.0020  &\textbf{.72$\pm$.0018} & .08$\pm$.0120           & \textbf{.68$\pm$.0208} & .21$\pm$.0163 & \textbf{.47$\pm$.0155} & .56$\pm$.0264 & \textbf{.96$\pm$.0072} \\
\textbf{KNN Dist $\downarrow$} &1772.50 &\textbf{1380.22} &126.61             & \textbf{72.54} &360.32  & \textbf{220.30} &139.09  & \textbf{123.07} \\
\textbf{FID $\downarrow$} &52.51 &\textbf{23.72} &8.95 & \textbf{0.45} &8.46  & \textbf{6.51} &1.69  & \textbf{1.32} \\ \bottomrule
\end{tabular}
}
\caption{Attack performance comparison on various datasets. $\uparrow$ and $\downarrow$ respectively symbolize that higher and lower scores give better attack performance.}
\label{tab: dataset}
\end{table*}

\paragraph{Cross-dataset experiment.} We study the effect of distribution shift between public and private data on the attack performance.
We train our GAN on Flickr-Faces-HQ Dataset (FFHQ)~\cite{karras2019style} and FaceScrub~\cite{ng2014data} to attack the target network VGG16 trained on CelebA. The attack results are presented in Table \ref{tab: cross}, which shows that both GMI and our attack suffer from a performance drop while ours still outperforms GMI. We notice that the performance drop on FaceSrub is larger than that on FFHQ. One possible reason is that images in FaceScrub have much lower resolution ($64\times64$), and there are a number of images under poor lighting conditions or only showing partial faces. This performance drop could potentially be resolved by using a GAN combined with unsupervised domain adaptation techniques and we will leave the exploration of this line of work to future work.

\begin{table}[ht!]
\centering
\resizebox{0.65\columnwidth}{!}{
\begin{tabular}{ccccc}
\toprule
\textbf{}     & \multicolumn{2}{c}{\textbf{FFHQ}$\rightarrow$\textbf{CelebA}}      & \multicolumn{2}{c}{\textbf{FaceScrub}$\rightarrow$\textbf{CelebA}} \\ 
\textbf{}     & GMI           & \textbf{Ours}          & GMI           & \textbf{Ours}          \\ \midrule
\textbf{Acc} $\uparrow$  & .15$\pm$.0015 & \textbf{.36$\pm$.0015} & .03$\pm$.0004     & \textbf{.13$\pm$.0008}     \\
\textbf{Acc5} $\uparrow$ & .35$\pm$.0017 & \textbf{.61$\pm$.0012} & .11$\pm$.0011     & \textbf{.30$\pm$.0015}   \\
\textbf{KNN Dist} $\downarrow$ &3014.45  &\textbf{2994.32}  &3003.90   &\textbf{2997.52}  \\
\textbf{FID} $\downarrow$ &69.12  &\textbf{36.02}  &112.83   &\textbf{60.05}  
\\ \bottomrule
\end{tabular}
}
\caption{Attack performance comparison where there is large distributional shift between public and private data. $A\rightarrow B$ represents the setting when the target network is trained on dataset $B$ and the GAN is trained on dataset $A$ to distill a generic prior for reconstructions. $\uparrow$ and $\downarrow$ respectively symbolize that higher and lower scores give better attack performance.}
\label{tab: cross}
\end{table}

\paragraph{Ablation study.} We proposed a couple of ideas to improve the GMI attack in~\cite{zhang2020secret}, including (1) soft-label discrimination (SD), which enables the discriminator to differentiate soft-labels produced by the target network, (2) entropy minimization (EM), which minimizes the prediction entropy of images produced by the generator, and (3) distributional recovery (DR), which explicitly models and estimates the private data distribution. Note that EM can only be combined with our SD. This is because a canonical discriminator only performs real vs. fake classification and minimizing the entropy of the prediction outputs in this case would not encourage the embedding to retain features that are more likely to appear in private data. We have shown that the combination of all these ideas can lead to significant attack performance improvement over the GMI. Here, we conduct an ablation study to investigate the improvement introduced by each individual idea as well as any reasonable combinations of these ideas. Table~\ref{tab: Ablation} presents the result of ablation study for attacking VGG16 trained on the CelebA dataset. We observe that both the attack accuracy and image quality get improved when we apply the idea of SD or DR. Adding entropy minimization can further improve the performance. The combination of the three ideas leads to the largest improvement.

\begin{table}[h!]
\centering
\resizebox{0.9\columnwidth}{!}{
\begin{tabular}{c|c|ccccc}
\toprule
                   & \textbf{GMI}               & \textbf{SD}            & \textbf{SD+EM}               & \textbf{DR}          & \textbf{SD+DR}        & \textbf{SD+EM+DR}      \\ \midrule
\textbf{Acc}       & .21$\pm$.0020 & .35$\pm$0042  & .43$\pm$.0035 & .47$\pm$.0022 & .62$\pm$.0028 & .72$\pm$.0018 \\
\textbf{Acc5}      & .43$\pm$.0014 & .60$\pm$.0013 & .68$\pm$.0017 & .74$\pm$.0024 & .87$\pm$.0003 & .92$\pm$.0003 \\
\textbf{KNN Dist}  & 1772.50                   &  1653.53                          & 1618.51           & 1562.48               & 1418.46                           &   1380.22    \\ 
\textbf{FID}       & 52.51                      & 33.75                      & 31.09           & 45.28           & 23.82                      & 23.72                      \\
\bottomrule
\end{tabular}
}
\caption{Ablation study of ideas introduced in this paper, including soft-label discrimination (SD), entropy minimization (EM), and distributional recovery (DR).}
\label{tab: Ablation}
\end{table}

\begin{table}[h!]
    \centering
    \tiny
    \resizebox{0.8\columnwidth}{!}{
    \begin{tabular}{c c c c c }
    \toprule
    & \multicolumn{2}{c}{\textbf{F\&I}} & \multicolumn{2}{c}{\textbf{F\&V }}\\
    &\textbf{GMI}&\textbf{Ours} &\textbf{GMI}&\textbf{Ours} \\
    \hline
    \textbf{Attack Acc} &.51$\pm$.0030&\textbf{.90$\pm$.0009}    &.51$\pm$.0048&\textbf{.90$\pm$.0005 }     \\
    \textbf{Top-5 Attack acc}
    &.78$\pm$.0025&\textbf{.99$\pm$.0001}   &.75$\pm$.0043&\textbf{.98$\pm$.0002   }  \\
    \textbf{KNN Dist} &1527.94&\textbf{1287.45} &1528.32&\textbf{1253.12} \\
    \textbf{FID}&54.89&\textbf{29.37} &54.76&\textbf{28.66}\\
    \hline
    &\multicolumn{2}{c}{\textbf{I\&V}} & \multicolumn{2}{c}{\textbf{F\&I\&V}}\\
    &\textbf{GMI}&\textbf{Ours} &\textbf{GMI}&\textbf{Ours}\\
    \textbf{Attack Acc} &.52$\pm$.0030&\textbf{.92$\pm$.0008 }  &.67$\pm$.0030&\textbf{.95$\pm$.0002}\\
    \textbf{Top-5 Attack acc}
    &.79$\pm$.0023&\textbf{.99$\pm$.0001 }  &.89$\pm$.0018&\textbf{1$\pm$0 }\\
    \textbf{KNN Dist}
    &1515.62&\textbf{1251.02} &1421.61&\textbf{1216.96}\\
    \textbf{FID}
     &54.80&\textbf{28.63} &53.73&\textbf{30.22}\\

    \bottomrule
    \end{tabular}
    }
    \caption{Attack performance on CelebA under multi-target setting. F, I, and V refer to face.evolve, IR152, and VGG16 respectively.}
    \label{tab:multi-target-results}
\end{table}

\paragraph{Extension to Multi-Target Model Inversion Attacks.}

So far, existing MI attack methods mainly focus on attacking a single target model. It is interesting to study attack performance when multiple different models trained on the same private dataset are available. %\textcolor{red}{add one scenario example}.
Will the attacker gain more information about this private dataset in this case? 
The proposed method can be easily extended to the multi-target MI attacks by combining the training losses over multiple target models. The details about the method are given in the supplementary material.

Table~\ref{tab:multi-target-results} shows the result of our method under the Multi-Target attack setting. We attack all possible combinations of the three target models used in the experiment shown in Table~\ref{tab: model}. 
% Exact same parameters are used as the single target threat model. We set $w_m$ in equations \ref{eq:L_supervised_MTA} and \ref{eq:L_iden_MTA} to be $\frac{1}{m}$ for all targets so that the discriminator supervised loss and the identity loss are averaged over all target models.

It is clear from Table~\ref{tab:multi-target-results} that the attack performance increases considerably under Multi-Target setting with both GMI and our approach. %Interestingly, attacking a target model with a weaker model can still provide preferable amount of additional information about the private dataset.
%For example, although IR152 or face.evolve have more complex model and higher attack accuracy (compared to VGG16 ) under single target threat model, 
% this paragraph needs to be restated in a more clear manner
For example, When IR152 or face.evolve are jointly utilized with VGG16, their attack accuracy increased by 9\% and 11\% receptively over their accuracies under single-target setting, even though VGG16 yields a weaker attack accuracy. 
Moreover, increasing the number of target models to three models further improved the attack performance. By attacking multiple target models jointly, our approach achieves an attacking accuracy of over $0.9$ in these experiments, which marks a significant milestone for multi-target model inversion attack.

\section{Conclusion}
In this paper, we propose several techniques that can significantly improve whitebox MI attacks against DNNs. Specifically, we propose to customize the training of a GAN to better distill knowledge useful for performing inversion attacks from public data. Additionally, we propose to build an explicit parameteric model for the private data distribution and present methods to estimate its parameters. Our experiments show that the combination of the proposed techniques can lead to the state-of-the-art attack performance on various datasets, models, and even when the public data has a large distributional shift from private data. We also extend our work to a new attack setting where multiple models trained on the same private dataset are available.
% We introduced a new threat model in which the attacker can utilize multiple targets that are trained on the dataset and achieve better attack performance.
For future work, we will investigate the potential application of these techniques to improve the MI attack in the blackbox setting. 

% \section*{Acknowledgments}
% This was was supported in part by......

%Bibliography
\bibliographystyle{unsrt}  
\bibliography{references}  

\clearpage
\appendix

\section*{Appendix}
\section{Extension to Multi-Target Model Inversion Attacks}
Existing MI attacks aim to extract sensitive features (or entirely construct samples) from a private dataset.
So far, all previous works on MI attacks only consider a single target model. However, multiple models trained on the same private dataset are often available in a real scenario.
% For example, there are BERT\_base and BERT\_large which are trained on \textcolor{red}{add examples here}
Performing MI attack on multiple different target models that are trained on the same private dataset could be an interesting and practical problem in real-world applications. Will the attacker gain more information about this private dataset in this case?

% We introduce this new setting ``Multi-Target Model Inversion Attacks'' where multiple target models that trained on the same private dataset, are available to the attacker. We show that a more powerful attack can be constructed by jointly attacking on the multiple target models.
%The intuition is that the attacker can learn exclusive features from different target models.

In our experiments, we extend our attack algorithm over single-target setting to the multi-target setting by following the same attacking procedure. First, an inversion-specific GAN is built under the guidance of multiple target models, where the supervised loss of discriminator $L_\text{multi\_sup}$ is a weighted combination of cross-entropy losses between the discriminator output distributions and soft labels given by different target models; supervised loss and generator loss are exactly same as under the single-target setting:
% \textcolor{red}{do we need to drop the loss equations that are same as under single setting?}
\begin{align}
    &L_\text{multi\_G} = \left\|\mathbb{E}_{x \sim p_\text{data}} \mathbf{f}(x)-\mathbb{E}_{z \sim \text{noise
    }} \mathbf{f}(G(z))\right\|_{2}^{2} + \lambda_h L_\text{entropy} \\
   &L_\text{multi\_D} = L_\text{multi\_sup} + L_\text{multi\_unsup} 
\end{align}
where
 \begin{align} \medmath{
    \label{eq:L_supervised_MTA}
     L_\text{multi\_sup}=-\mathbb{E}_{x \sim p_{\text{data }}(x)}\sum_{m=1}^M w_{m} \sum_{k=1}^K T_k^m(x) \log p_{\text{disc}}(y=k \mid x)
     }
 \end{align}
and
\begin{align}
   L_\text{multi\_unsup}=-\{&\mathbb{E}_{x \sim p_{\text{data }}(x)} \log D(x)+ \\
   &\mathbb{E}_{z \sim \text{noise}} \log (1-D(G(z)))\}
\end{align}

Here we use the same notation as the single-target setting. $M$ is the number of target models, $w_m$ is the weight on the loss over the $m$th target model $T^m$.
%$T_k^m$ denotes the output probability of a generated sample being assigned to the given target label $k$ under target model $T^m$.
 
Then the learned GAN over multiple target models are applied for distributional recovery where the new identity loss is a weighted combination of identity loss over multiple target models, and prior loss remain unchanged:
 \begin{align}
     \label{eq:L_iden_MTA}
     &L_\text{multi\_id} =-\sum_{m=1}^M \frac{w_{m}}{L}\sum_{l=1}^L\log{T_k^m(G(\sigma\epsilon_l+\mu))}\\ 
     &L_\text{multi\_prior} = -\mathbb{E}_{z'\sim p_\text{gen}}\log D(G(z'))\\
 \end{align}
and overall loss for distributional recovery under multi-attack setting is:
\begin{align}
    L_\text{multi} = L_\text{multi\_prior}+ \lambda_i L_\text{multi\_id}
\end{align}

In our experiments, the same set of training hyperparameters such as learning rate and weight decay are used.
In addition, we set $w_m$ in  (\ref{eq:L_supervised_MTA}) and (\ref{eq:L_iden_MTA}) to $\frac{1}{M}$ so that the discriminator supervised loss and the identity loss are averaged over all target models.

\end{document}